\documentclass{article}
\usepackage[font={small,it}]{caption}
\usepackage[authoryear,compress]{natbib}
\usepackage{soul, color}
\usepackage{colortbl}
\usepackage{bm}
\usepackage{graphicx}
\usepackage{booktabs}
\usepackage{caption}
\usepackage{arydshln}
\usepackage{subcaption}
\usepackage{amsmath}
\usepackage{amssymb}
\usepackage{amsthm}
\usepackage{dsfont}
\usepackage{bbold}
\usepackage{listings}
\usepackage{setspace}
\usepackage{url}
\usepackage{verbatim}
\usepackage{algorithm}
\usepackage{algpseudocode}
\usepackage{nicefrac}
\usepackage{mwe}
\usepackage{makecell}
\usepackage{multirow}
\usepackage{tikz}
\usetikzlibrary{positioning,decorations.pathreplacing,shapes}
\newcommand{\E}{\mathbb{E}}

\newcommand{\Sbf}{\mathbf{S}}
\newcommand{\Gbf}{\mathbf{G}}
\newcommand{\Ibf}{\mathbf{I}}
\newcommand{\Abf}{\mathbf{A}}
\newcommand{\BH}{\hat{\mathbf{B}}}
\newcommand{\UH}{\hat{\mathbf{U}}}
\newcommand{\YH}{\hat{\mathbf{Y}}}
\newcommand{\Wbf}{\mathbf{W}}
\newcommand{\Hbf}{\mathbf{H}}
\newcommand{\Zbf}{\mathbf{Z}}
\newcommand{\Scal}{\mathcal{S}}
\newcommand{\bbf}{\mathbf{b}}
\newcommand{\ubf}{\mathbf{u}}
\newcommand{\ybf}{\mathbf{y}}
\newcommand{\xbf}{\mathbf{x}}
\newcommand{\zbf}{\mathbf{z}}
\newcommand{\sbf}{\mathbf{s}}

\newcommand{\rr}{\mathbb{R}}
\newcommand{\pp}{\mathcal{P}}
\newcommand{\kl}{KL}

\newcommand{\nutil}{\tilde{\nu}}
\newcommand{\pitil}{\tilde{\pi}}
\newcommand{\nuhat}{\hat{\nu}}
\newcommand{\pihat}{\hat{\pi}}

\newcommand{\supp}{supp}
\newcommand{\resample}{\textbf{Resample}}
\newcommand{\Unif}{Unif}
\newcommand{\ADI}{ADI}

\definecolor{codegreen}{rgb}{0,0.6,0}
\definecolor{codegray}{rgb}{0.5,0.5,0.5}
\definecolor{codepurple}{rgb}{0.58,0,0.82}
\definecolor{backcolour}{rgb}{0.95,0.95,0.92}
\definecolor{Gray}{gray}{0.8}
\definecolor{lightyellow}{RGB}{245,238,197}
\definecolor{lightblue}{RGB}{204, 230, 255}
\newcolumntype{a}{>{\columncolor{lightyellow}}c}
\newcolumntype{b}{>{\columncolor{backcolour}}c}

\lstdefinestyle{mystyle}{
    backgroundcolor=\color{backcolour},
    commentstyle=\color{codegreen},
    keywordstyle=\color{magenta},
    numberstyle=\tiny\color{codegray},
    stringstyle=\color{codepurple},
    basicstyle=\ttfamily\footnotesize,
    breakatwhitespace=false,
    breaklines=true,
    captionpos=b,
    keepspaces=true,
    numbers=left,
    numbersep=5pt,
    showspaces=false,
    showstringspaces=false,
    showtabs=false,
    tabsize=2
}

\DeclareFontFamily{U}{mathx}{}
\DeclareFontShape{U}{mathx}{m}{n}{<-> mathx10}{}
\DeclareSymbolFont{mathx}{U}{mathx}{m}{n}
\DeclareMathAccent{\widehat}{0}{mathx}{"70}
\DeclareMathAccent{\widecheck}{0}{mathx}{"71}

\newtheorem{assumption}{Assumption}
\newtheorem{definition}{Definition}
\newtheorem{proposition}{Proposition}

\definecolor{codegreen}{rgb}{0,0.6,0}

\newcommand{\keywords}[1]{\noindent \textbf{Keywords:}\; #1}

\begin{document}

\title{Efficient probabilistic reconciliation of forecasts for real-valued and count time series}
\author{Lorenzo Zambon\thanks{IDSIA, Dalle Molle Institute for Artificial Intelligence, CH-6962, Lugano, Switzerland; \texttt{\{lorenzo.zambon, dario.azzimonti, giorgio.corani\}@idsia.ch}} $^{,} \hspace{-1.mm}$
\thanks{Department of Mathematics, University of Pavia, 27100, Pavia, Italy}
\and Dario Azzimonti\footnotemark[1]
\and Giorgio Corani\footnotemark[1]}
\date{}

\maketitle

\abstract{
Hierarchical time series are common in several applied fields. 
The forecasts for these time series are required to be coherent, that is, to satisfy the constraints given by the hierarchy. 
The most popular technique to enforce coherence is called reconciliation, which adjusts the base forecasts computed for each time series.
However, recent works on probabilistic reconciliation present several limitations. 
In this paper, we propose a new approach based on conditioning to reconcile any type of forecast distribution.
We then introduce a new algorithm, called Bottom-Up Importance Sampling, to efficiently sample from the reconciled distribution.
It can be used 
for any base forecast distribution: discrete, continuous, or in the form of samples,  providing a major speedup compared to the current methods.
Experiments on several temporal hierarchies show a significant improvement over base probabilistic forecasts.
}

\bigskip

\keywords{Forecast reconciliation, Probabilistic reconciliation, Temporal hierarchies, Importance sampling}

\section{Introduction}
Often time series are organized into a hierarchy. For example, the total visitors of a country can be divided into regions and the visitors of each region can be further divided into sub-regions.
Such data structures are referred to as \textit{hierarchical time series}; they are common in  fields such as retail sales \citep{makridakis2021m5} and energy modelling \citep{taieb2021hierarchical}.

The forecasts for hierarchical time series should 
respect some summing constraints, in which case they are referred to as \textit{coherent}.
For instance, the sum of the forecasts for the sub-regions should match the forecast for the entire region.
However, the forecasts   independently produced for each time series 
(\textit{base forecasts}) are generally \textit{incoherent}.

\textit{Reconciliation} algorithms \citep{hyndman2011optimal, Wickramasuriya.etal2018} adjust the incoherent base forecasts, 
making them coherent. 
Reconciled forecasts are generally more accurate than base forecasts: indeed,  forecast reconciliation is 
a special case  
of forecast combination \citep{hollyman2021understanding}.
An important application of reconciliation
algorithms
is constituted by temporal hierarchies \citep{athanasopoulos2017_temporal, KOURENTZES-elucidate}, which  make coherent
the forecasts  produced for the same time series at different temporal scales.

Most reconciliation algorithms \citep{hyndman2011optimal, Wickramasuriya.etal2018, wickramasuriya2020optimal, difonzo2021cross, difonzo2022forecast}
provide only  \textit{reconciled point forecasts}.
It is however clear \citep{kolassa_we_2023} that   \textit{reconciled predictive distributions} 
are needed for decision making.

Probabilistic reconciliation has been addressed only recently; earlier attempts 
\citep{jeon2019probabilistic, taieb2021hierarchical},
though experimentally effective,
lacked a strong formal justification.
For the case of Gaussian base forecasts,
    \cite{corani_ecml_reconc}  obtains the reconciled distribution in analytical form introducing the approach of \textit{reconciliation via conditioning}.
\cite{panagiotelis2022probabilistic}  provides
a  framework for
probabilistic reconciliation via projection.
However this approach cannot reconcile discrete distributions.
%
\cite{corani2022probabilistic} 
performs probabilistic reconciliation via conditioning of count time series
by adopting the concept of \textit{virtual evidence} \citep{pearl1988probabilistic}.
However its implementation in probabilistic programming, based on Markov Chain Monte Carlo (MCMC), is too slow on large hierarchies;  moreover it requires the base forecast distribution to be in parametric form.

The main contribution of this paper is the Bottom-Up Importance Sampling (BUIS) algorithm, which  samples from the reconciled distribution obtained via conditioning with a substantial speedup with respect to  \citet{corani2022probabilistic}.
BUIS can be used even when the base forecast distribution is only available through samples.
This is the case of forecasts returned by models for time series  of counts
\citep{tscount} or  based on deep learning \citep{salinas2020deepar}.
We prove the convergence of BUIS to the actual reconciled distribution. An implementation of the algorithm in the \textit{R} language is available  in  the \textit{R} package \textit{bayesRecon} \citep{bayesRecon}.

We provide two further formal contributions. The first is a definition of coherence for probabilistic forecasts that applies to both discrete and continuous distributions. 
The second is a novel interpretation of the reconciliation via conditioning,
in which the base forecast distribution is conditioned on the hierarchy  constraints. 
This allows for a unified treatment of the reconciliation of discrete and continuous forecast distributions.
We test our method exhaustively on temporal hierarchies  reporting positive results both for the accuracy and the efficiency of our method.

The paper is organized as follows. 
In Sec.~\ref{sec: notation}, we introduce the notation and  the reconciliation of point forecasts.
In Sec.~\ref{sec: probabilistic reconciliation}, we introduce
our approach to reconciliation via conditioning and we compare it to the existing literature.
In Sec.~\ref{sec: Sampling from the reconciled distribution},
we introduce the  Bottom-Up Importance Sampling algorithm.
We empirically verify its correctness in Sec.~\ref{sec: Experiments on synthetic data}, while in Sec.~\ref{sec: Experiments} we test it on different data sets.
We present the conclusions  in Sec.~\ref{sec: conclusion}.

\section{Notation}\label{sec: notation}

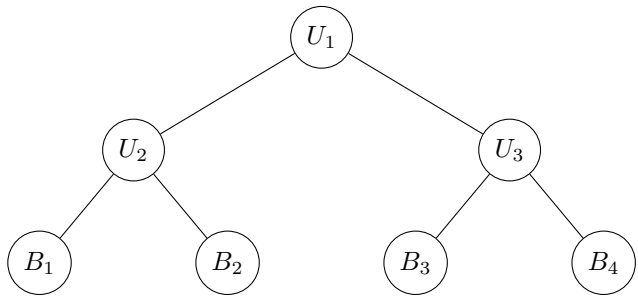
\begin{figure*}
    \centering
\begin{tikzpicture}[level/.style={sibling distance=50mm/#1}]
\node [circle,draw] {$U_1$}
  child {node [circle,draw]  {$U_2$}
    child {node [circle,draw] {$B_1$}
    }
    child {node [circle,draw] {$B_2$}
    }
  }
  child {node [circle,draw]  {$U_3$}
    child {node [circle,draw]{$B_3$}
    }
  child {node [circle,draw] {$B_4$}
  }
};
\end{tikzpicture}
    \vspace*{2 mm}
    \caption{A hierarchy with $4$ bottom and $3$ upper variables.}
    \label{fig: simple tree}
\end{figure*}

Consider the  hierarchy of  Fig.~\ref{fig: simple tree}.
We denote by $\bbf = [b_1,\dots,b_{n_b}]^T$ the vector of bottom variables, and by $\ubf = [u_1,\dots,u_{n_u}]^T$ the vector of upper variables. 
We then denote by 
\[
\ybf = \begin{bmatrix}
          \ubf \\
          \bbf
         \end{bmatrix} \in \rr^n
\]
the vector of all the variables.
The hierarchy can be expressed as a set of linear constraints:
\begin{equation}\label{eq: def S}
\ybf = \Sbf \bbf, \; \text{where} \; \Sbf= \begin{bmatrix}
          \Abf \\ \hdashline[2pt/2pt]
          \Ibf
         \end{bmatrix}.
\end{equation}
We refer to $\Ibf \in \rr^{n_b \times n_b}$ as the \textit{identity matrix}, to
$\Sbf \in \rr^{n \times n_b}$ as the \textit{summing matrix} and to $\Abf \in \rr^{n_u \times n_b}$ as the \textit{aggregating matrix}. 
We can thus write the constraints as $\ubf = \Abf \bbf$.
For example, the aggregating matrix of the hierarchy in Fig.~\ref{fig: simple tree} is: 
\[
A = \begin{bmatrix}
          1 & 1 & 1 & 1 \\
          1 & 1 & 0 & 0 \\
          0 & 0 & 1 & 1
         \end{bmatrix}.
\]
A point $\ybf \in \rr^n$ is \textit{coherent} if it satisfies the constraints given by the hierarchy. 
We denote by $\Scal$ the set of coherent points, 
which is a linear subspace of $\rr^n$:
\begin{equation}\label{eq: def S cal}
\mathcal{S} := \{ \ybf \in \rr^n : \; \ybf = \Sbf \bbf \}.    
\end{equation}

\subsection{Temporal hierarchies}
In temporal hierarchies \citep{athanasopoulos2017_temporal, KOURENTZES-elucidate}, 
forecasts are generated for the same time series at different temporal scales. 
For instance, a quarterly time series can be aggregated to the semi-annual and the annual scale.
If we are interested in predictions up to one year ahead, we compute  four quarterly forecasts $\hat{q}_1, \hat{q}_2, \hat{q}_3, \hat{q}_4$,
two semi-annual forecasts $\hat{s}_1,\hat{s}_2$, and an annual forecast $\hat{a}_1$.
We then obtain the hierarchy in Fig.~\ref{fig: simple tree}. 
The base point forecasts,  independently computed at each frequency, are $\hat{\bbf} = [\hat{q}_1, \hat{q}_2, \hat{q}_3, \hat{q}_4]^T$ and $\hat{\ubf} = [\hat{a}_1, \hat{s}_1, \hat{s}_2]^T$.

\subsection{Point forecasts reconciliation}

Let us  denote by $\hat{\ybf} = \big[\hat{\ubf}^T\,\vert\,\hat{\bbf}^T\big]^T$ the vector of the base (incoherent) forecasts. 
Note that, for ease of notation, we drop the time subscript.
Point reconciliation is generally performed in two steps \citep{hyndman2011optimal, Wickramasuriya.etal2018}. 
First, the reconciled bottom forecasts are computed by linearly combining the base forecasts of the entire hierarchy:
\[
\Tilde{\bbf} = \Gbf \hat{\ybf},
\]
for some matrix $\Gbf \in \rr^{m\times n}$.
Then, the reconciled forecasts for the whole hierarchy are given by:
\[
\Tilde{\ybf} = \Sbf \Tilde{\bbf}.
\]
The state-of-the-art reconciliation method is
MinT \citep{Wickramasuriya.etal2018}, which defines $\Gbf$ as:
\[
\Gbf = (\Sbf^T \Wbf^{-1} \Sbf)^{-1} \Sbf^T \Wbf^{-1},
\]
where $\Wbf$ is the covariance matrix of the errors of the base forecasts.
This method minimizes  the 
expected sum of the squared errors 
of the reconciled forecasts, under the assumption of unbiased base forecasts.

\subsection{Probabilistic reconciliation}
Probabilistic reconciliation   requires a probabilistic framework, in which forecasts are in the form of probability distributions. 
We denote by $\hat{\nu} \in \pp(\rr^n)$ the forecast distribution for $\ybf$, where $\pp(\rr^n)$ is the space of probability measures on $\left(\rr^n, \mathcal{B}(\rr^n)\right)$, 
and $\mathcal{B}(\rr^n)$ is the Borel $\sigma$-algebra on $\rr^n$.
Moreover, we denote by $\hat{\nu}_u$ and $\hat{\nu}_b$ the marginal distributions of, respectively, the forecasts for the upper and the bottom components of $\ybf$.

The forecast distribution $\hat{\nu}$ may be either discrete or absolutely continuous.
In the following, if there is no ambiguity, we will use $\hat{\pi}$ to denote either its probability mass function, in the former case, or its density, in the latter.
Therefore, if $\hat{\nu}$ is discrete, we have
\[
\hat{\nu}(F) = \sum_{x \in F} \hat{\pi}(x),
\]
for any $F \in \mathcal{B}(\rr^n)$.
Note that the sum is well-defined as $\hat{\pi}(x) > 0$ for at most countably many $x$'s. 
On the contrary, if $\hat{\nu}$ is absolutely continuous, for any $F \in \mathcal{B}(\rr^n)$ we have
\[
\hat{\nu}(F) = \int_F \hat{\pi}(x) \, dx.
\]


\section{Probabilistic Reconciliation}
\label{sec: probabilistic reconciliation}
We now discuss coherence in the probabilistic framework and  our approach to probabilistic reconciliation.

Recall that a point forecast is incoherent if it does not belong to the set $\mathcal{S}$, defined as in \eqref{eq: def S cal}. 
Let $\hat{\nu} \in \pp(\rr^n)$ be a forecast distribution. 
Thus, $\hat{\nu}$ is incoherent if there exists a set $T$ of incoherent points, i.e. $T \cap \mathcal{S} = \emptyset$, 
such that $\hat{\nu}(T) > 0$.
Or, equivalently, if $\supp(\hat{\nu}) \nsubseteq \mathcal{S}$.
We now define the summing map $s: \rr^{n_b} \to \rr^n$ as 
\begin{equation} \label{eq: def map s}
s(\mathbf{b}) = \Sbf \mathbf{b}.    
\end{equation}
The image of $s$ is given by $\mathcal{S}$. 
Moreover, from \eqref{eq: def map s} and \eqref{eq: def S}, $s$ is injective. 
Hence, $s$ is a bijective map between $\rr^{n_b}$ and $\mathcal{S}$, with inverse given by $s^{-1}(\ybf) = \bbf$, where $\ybf = (\ubf, \bbf) \in \mathcal{S}$. 
As explained in \cite{panagiotelis2022probabilistic}, for any $\nu \in \pp(\rr^{n_b})$ we may obtain a distribution $\nutil \in \pp(\mathcal{S})$ as $\nutil = s_{\#} \nu$, namely the pushforward of $\nu$ using $s$:
\[
\nutil(F) = \nu(s^{-1}(F)), \qquad \forall \, F \in \mathcal{B}(\mathcal{S}),
\]
where $s^{-1}(F) := \{ \mathbf{b} \in \rr^{n_b} : s(\mathbf{b}) \in F \}$ is the preimage of $F$. 
In other words, $s_{\#}$ builds a probability distribution for $\ybf$ supported on the coherent subspace $\mathcal{S}$ from a distribution on the bottom variables $\bbf$.
Since $s$ is a measurable bijective map, $s_{\#}$ is a bijection between $\pp(\rr^{n_b})$ and $\pp(\mathcal{S})$, with inverse given by $(s^{-1})_{\#}$ (Appendix \ref{sec: appendix proofs}).
We thus propose the following definition.
\begin{definition}\label{def: coherent distribution}
We call coherent distribution any distribution $\nu \in \pp(\rr^{n_b})$.
\end{definition}
This  definition works with any type of distribution. 
Moreover, it can be used even if the constraints are not linear, as it does not require $s$ to be a linear map.

\subsection{Probabilistic reconciliation}

The aim of probabilistic  reconciliation is to obtain a coherent reconciled distribution $\nutil \in \pp(\rr^{n_b})$ from the base forecast distribution $\hat{\nu} \in \pp(\rr^n)$.


The probabilistic bottom-up approach,
which simply ignores any probabilistic information about the upper series, is obtained by setting $\nutil = \hat{\nu}_b$.

\cite{panagiotelis2022probabilistic} proposes a reconciliation method based on projection. 
Given a continuous map $\psi: \rr^n \to \mathcal{S}$, the reconciled distribution $\tilde{\nu} \in \pp(\mathcal{S})$ is defined as the push-forward of the base forecast distribution  $\nuhat$ using $\psi$:
\[
\tilde{\nu} = \psi_{\#} \hat{\nu},
\]
i.e. $\tilde{\nu}(F) = \hat{\nu}(\psi^{-1}(F))$, for any $F \in \mathcal{B}(\rr^n)$.
Hence, if $\ybf_1, \dots, \ybf_N$ are independent samples from $\hat{\nu}$, 
then $\psi(\ybf_1), \dots, \psi(\ybf_N)$ are independent samples from $\nutil$.
The map $\psi$ is expressed as $\psi = s \circ g$, where
$g: \rr^n \to \rr^{n_b}$ combines information from all the levels by projecting on the bottom level.
$g$ is assumed to be in the form $g(\ybf) = \mathbf{d} + \Gbf\ybf$, and the parameters $\mathbf{\gamma} := (\mathbf{d}, \textit{vec}(\Gbf)) \in \rr^{n_b + n_b \times n}$ are optimized through stochastic gradient descent (SGD) to minimize a chosen scoring rule.
This approach therefore can only be used with continuous distributions.

\subsection{Probabilistic reconciliation via conditioning}

We now present our approach to probabilistic reconciliation, based on conditioning on the hierarchy constraints.
Let $\YH = (\UH,\BH)$ be a random vector representing the probabilistic forecasts with distribution
given by $\hat{\nu}$,
so that $\hat{\nu}_u$ and $\hat{\nu}_b$ are the distributions 
of  $\UH$ and $\BH$. 

Let us first suppose that the base forecast distribution $\hat{\nu} \in \pp(\rr^n)$ is discrete,
and let $\hat{\pi}$ be its probability mass function. 
We define $\nutil$ by conditioning on the coherent subspace $\mathcal{S}$:
\begin{align}\label{eq: mutil discrete}
\nutil(F) 
&= \mathbb{P}(\BH \in F \mid \YH \in \mathcal{S}) \nonumber \\
&= \frac{ \mathbb{P}(\BH \in F,\, \YH \in \mathcal{S}) }
{\mathbb{P}(\YH \in \mathcal{S})} \nonumber \\
&= \frac{ \mathbb{P}(\BH \in F,\, \UH = \Abf \BH) }
{\mathbb{P}(\UH = \Abf \BH)} \nonumber \\
&= \frac{\sum_{\bbf \in F} \hat{\pi}(\Abf \bbf, \bbf)}{\sum_{\xbf \in \rr^{n_b}} \hat{\pi}(\Abf \xbf, \xbf)},
\end{align}
for any $F \in \mathcal{B}(\rr^{n_b})$, provided that $\mathbb{P}(\YH \in \mathcal{S}) > 0$.
The sums in \eqref{eq: mutil discrete} are well-defined, as $\hat{\pi}(\ubf,\bbf)=\hat{\pi}(\ybf) > 0$ for at most countably many $\ybf$'s.
Hence, $\nutil$ is a discrete probability distribution with pmf given by 
\begin{equation}\label{eq: mutil pmf}
\Tilde{\pi}(\bbf)
= \frac{\hat{\pi}(\Abf \bbf, \bbf)}{\sum_{\xbf \in \rr^{n_b}} \hat{\pi}(\Abf \xbf, \xbf)} 
\propto \hat{\pi}(\Abf \bbf,\bbf).
\end{equation}
%
Note that, if $\hat{\nu}$ is absolutely continuous, we have that $\hat{\nu}(\mathcal{S}) = 0$, since the Lebesgue measure of $\mathcal{S}$ is zero. 
Hence, $\mathbb{P}(\BH \in F \mid \YH \in \mathcal{S})$ is not well-defined. 
However, 
if we denote by $\hat{\pi}$ the density of $\hat{\nu}$, 
the last expression is still well-posed.
We thus give the following definition.
%
%
%
\begin{definition}\label{def: reconciled distribution}
Let $\hat{\nu} \in \pp(\rr^n)$ be a base forecast distribution.
The reconciled distribution through conditioning is defined as the probability distribution $\tilde{\nu} \in \pp(\rr^{n_b})$ such that 
\begin{equation}\label{eq: def mutil}
\Tilde{\pi}(\bbf)
\propto \hat{\pi}(\Abf \bbf,\bbf),
\end{equation}
where $\hat{\pi}$ and $\tilde{\pi}$ are the densities of (respectively) $\hat{\nu}$ and $\tilde{\nu}$, if $\hat{\nu}$ is absolutely continuous, or the probability mass functions otherwise.
\end{definition}
To rigorously derive \eqref{eq: def mutil} in the continuous case, we proceed as follows.
Let us define the random vector $\Zbf:= \UH - \Abf \BH$. 
Note that the event $\{\YH \in \mathcal{S}\}$ coincides with $\{\Zbf = \textbf{0}\}$.
The joint density of $(\Zbf,\BH)$ can be easily computed (Appendix \ref{sec: appendix proofs}):
\[
\pi_{(\Zbf,\BH)}(\zbf,\bbf) = \hat{\pi}(\zbf + \Abf \bbf, \bbf).
\]
Then, the conditional density of $\BH$ given $\Zbf = \textbf{0}$ is given by
\cite[Chapter~4]{cinlar2011probability}:
\begin{align}
\Tilde{\pi}(\bbf) 
&= \frac{\pi_{(Z,B)}(\textbf{0},\bbf)}{\int_{\rr^{n_b}} \pi_{(Z,B)}(\textbf{0},\xbf) \, d\xbf} \nonumber\\
&= \frac{\hat{\pi}(\Abf \bbf, \bbf)}{\int_{\rr^{n_b}} \hat{\pi}(\Abf \xbf, \xbf) \, d\xbf} \nonumber \\
&\propto \hat{\pi}(\Abf \bbf, \bbf), \nonumber
\end{align}
provided that $\int_{\rr^{n_b}} \hat{\pi}(\Abf \xbf, \xbf) \, d\xbf > 0$.
Finally, note that, if $\UH$ and $\BH$ are independent, \eqref{eq: def mutil} may be rewritten as
\begin{equation}\label{eq: mutil indep}
\Tilde{\pi}(\bbf) \propto \hat{\pi}_u(\Abf \bbf) \cdot \hat{\pi}_b(\bbf),
\end{equation}
where $\hat{\pi}_u$ and $\hat{\pi}_b$ are the densities of (respectively) $\hat{\nu}_u$ and $\hat{\nu}_b$.
This approach can be applied to both continuous and discrete distributions, yielding the same expression \eqref{eq: def mutil} for the reconciled distribution.

Given two coherent points $\ybf_1, \ybf_2 \in \mathcal{S}$,
the distribution reconciled  through conditioning satisfies the following property:
\begin{equation}\label{eq: odds ratio}
\frac{\Tilde{\pi}(\ybf_1)}{\Tilde{\pi}(\ybf_2)}
= \frac{\hat{\pi}(\ybf_1)}{\hat{\pi}(\ybf_2)}
\end{equation}
if $\hat{\pi}(\ybf_2) \neq 0$, and $\Tilde{\pi}(\ybf_2) = 0$ 
if $\hat{\pi}(\ybf_2) = 0$;
i.e., the relative probabilities of the coherent points are preserved. 
%
Moreover, reconciliation via conditioning  ignores the behaviour of the base distribution outside the coherent subspace.
As shown by \eqref{eq: def mutil}, $\nutil$ only depends on the values of $\hat{\nu}$ on $\mathcal{S}$. 
Reconciliation via conditioning is therefore invariant under modifications of the base forecast probabilities outside the coherent subspace. 
This constitutes a major difference with respect to the method of
\cite{panagiotelis2022probabilistic}
that will be thoroughly studied in future work.

In \cite{corani2022probabilistic}, the authors follow an approach based on 
\textit{virtual evidence} \citep{pearl1988probabilistic} to reconcile discrete forecasts. 
They set the joint bottom-up distribution as a prior on the entire hierarchy, and the update is made by conditioning on the base upper forecasts, treated  as uncertain observations.
In contrast, 
we provide a unified treatment of reconciliation via conditioning for the discrete and the continuous case.
Our approach has a clear interpretation, as the conditioning is done on the hierarchy constraints.

\section{Sampling from the reconciled distribution}
\label{sec: Sampling from the reconciled distribution}

If the base forecasts are jointly Gaussian, then the reconciled distribution is also Gaussian.
In this case, reconciliation via conditioning yields
the same mean and variance  \citep{corani_ecml_reconc}  of
MinT, which is optimal with respect to
the log score \citep{wickramasuriya2021probabilistic}.

In general, however, the reconciled distribution is not available in parametric form, hence we need to resort to sampling approaches. 
We propose a  method  based on
Importance Sampling (IS, \cite{kahn1950random, elvira2021advances}).

\subsection{Importance Sampling}
\label{sec: importance sampling}

Let $X$ be an absolutely continuous random variable with density $p$.
Suppose we want to compute the expectation $\mu = \E[f(X)]$, for some function $f$.
Importance Sampling  estimates the expectation $\mu$ by sampling from a different distribution $q$, and by weighting the samples to correct  the mismatch between the \textit{target} $p$ and the \textit{proposal} $q$. 

In the following the term \textit{density}  denotes either the probability mass function (for discrete distributions) or the density with respect to the Lebesgue measure (for absolutely continuous distributions).
Let $q$ be a density such that $q(x)>0$ if $f(x) p(x) \neq 0$, and let $y_1,\dots,y_N$ be independent samples drawn from $q$.
The self-normalized importance sampling estimate \citep{elvira2021advances} is:
\begin{equation}\label{eq: MC self-normalized IS estimate}
\E[f(X)] \approx \frac{\sum_{i=1}^N w(y_i) f(y_i)}{\sum_{i=1}^N w(y_i)},
\end{equation}
where $w$ is defined as $w(y) = c \, \frac{p(y)}{q(y)}$, for some (typically unknown) constant $c$.
%

\subsection{Probabilistic reconciliation via IS}
\label{sec: IS with mutil}

Let $\nutil$ (Definition \ref{def: reconciled distribution}) be the target distribution.
We set $\hat{\nu}_b$ as proposal distribution. 
Given a sample $\bbf_1,\dots,\bbf_N$ drawn form $\hat{\nu}_b$, the weights are computed as
\begin{equation}\label{eq: weights}
w_i := \frac{\hat{\pi}(\Abf \bbf_i, \bbf_i)}{\hat{\pi}_b(\bbf_i)}.
\end{equation}
Then, $(\bbf_i, \Tilde{w}_i)_{i=1,\dots,N}$ is a weighted sample from $\nutil$, where $\Tilde{w}_i := \nicefrac{w_i}{\sum_{j=1}^N w_j}$ are the normalized weights.
Note that \eqref{eq: weights} may be interpreted as the conditional density of $\UH$ at the point $\Abf \bbf_i$, given that $\BH = \bbf_i$.
We thus draw samples $(\bbf_i)_i$ from the base bottom distributions, and then weight how likely they are using the base upper distributions.
%
Under the assumption of independence between $\BH$ and $\UH$, the density of $\nutil$ factorizes as in \eqref{eq: mutil indep}, hence:
\begin{equation}
w_i = \hat{\pi}_u(\Abf \bbf_i).
\end{equation}

However, IS is affected by
the curse of dimensionality \citep{agapiou2017importance}.
In Appendix~\ref{sec: app large hier}, we empirically show that IS has poor accuracy when reconciling  large hierarchies.
Another shortcoming of IS is that it is  unreliable if the proposal distribution does not well approximate the target distribution.
Indeed, we prove in Appendix~\ref{sec: app KL} that the performance of IS degrades as the Kullback-Leibler divergence between bottom-up and base forecast distributions (which is  related to the incoherence of the base forecasts) increases. 
The Bottom-Up Importance Sampling (BUIS) algorithm addresses such problems.

\subsection{Bottom-Up Importance Sampling algorithm}
\label{sec: buis algorithm}

\begin{algorithm*}[!htp]
    \caption{Bottom-Up Importance Sampling}
    \label{alg:BUIS}
\begin{algorithmic}[1]
\State \textbf{Sample} $\big(\bbf^{(i)}\big)_{i=1,\dots,N}$ from $\hat{\pi}_b$
\For {$l$ in levels}
    \For {$j=1,\dots,k_l$}
        \State {$\widecheck{w}^{(i)} \gets \hat{\pi}_{u_{j, l}}\left(\sum_{t=1}^{q_{j,l}} b_{t, (j,l)}^{(i)}\right)$ 
        \quad for $i=1,\dots,N$} 
        \State $w^{(i)} \gets \frac{\widecheck{w}^{(i)}}{\sum_h \widecheck{w}^{(h)}}$ \quad for $i=1,\dots,N$
        \State  $\big(\Bar{\bbf}_j^{(i)}\big)_i \gets \resample
        \left( \left(b^{(i)}_{1, (j,l)}, \dots, b^{(i)}_{q_{j,l}, (j,l)}  \right), w^{(i)}\right)_i $ 
    \EndFor 
    \State $\bbf^{(i)} \gets \left[\Bar{\bbf}_1^{(i)},\dots,\Bar{\bbf}_{k_l}^{(i)}\right]$ \quad for $i=1,\dots,N$
\EndFor
\State \Return $\big(\bbf^{(i)}\big)_i$
\end{algorithmic}
\end{algorithm*}

First, we state the main assumption of our algorithm:
\begin{assumption}\label{assumptions: independent}
The base forecasts of each variable
are conditionally independent, given the time series observations.
\end{assumption}
We leave for future work the extension of this algorithm to deal with correlations between the base forecasts.
In this paper we perform experiments with  temporal hierarchies, which commonly make this assumption.

In order to simplify the presentation, we also  assume that the  data structure is strictly \textit{hierarchical}, i.e.,  that every node only has one parent and thus the hierarchy is represented by a tree. 
\textit{Grouped} time series \citep[Chapter~11]{hyndman2021forecasting}, which do not satisfy this assumption, require a more complex treatment; we discuss it in Sect.~\ref{sec: grouped time series}.

The BUIS algorithm exploits the hierarchical structure to split a large $n_u$-dimensional importance sampling problem into $n_u$ one-dimensional problems, thus deeply alleviating the curse of dimensionality.
BUIS  starts by drawing a sample from the base bottom distribution $\nuhat_b$. 
Then, for each level of the hierarchy, from bottom to top, it updates the sample through an importance sampling step, using the ``partially'' reconciled distribution  as proposal.

For each level $l=1,\dots,L$ of the hierarchy, we denote the upper variables at level $l$ by $u_{1,l}, \dots,u_{k_l, l}$. Moreover, for any upper variable $u_{j, l}$, we denote by $b_{1, (j,l)}, \dots, b_{q_{j,l}, (j,l)}$ the bottom variables that sum up to $u_{j, l}$. 
In this way, we have that $\sum_{l=1}^L k_l = n_u$, the number of upper variables, while $\sum_{j=1}^{k_l} q_{j,l} = n_b$, the number of bottom variables, for each level $l$. 

Let us consider, for example, the hierarchy in Fig.~\ref{fig: simple tree}.
For the first level $l=1$, we have $k_1 = 2$, $u_{1,1} = U_2$, and $u_{2,1} = U_3$. Moreover, $q_{1,1} = q_{2,1} = 2$, and $b_{1, (1,1)} = B_1$, $b_{2, (1,1)} = B_2$, $b_{1, (2,1)} = B_3$, $b_{2, (2,1)} = B_4$.
For the last level $l=2$, we have $k_2 = 1$, $u_{1,2} = U_1$, $q_{1,2} = 4$, $b_{1, (1,2)} = B_1$, $b_{2, (1,2)} = B_2$, $b_{3, (1,2)} = B_3$, $b_{4, (1,2)} = B_4$.

Alg.~\ref{alg:BUIS} shows the BUIS algorithm.
The ``Resample'' step 
 samples with replacement from the discrete distribution given by 
\begin{equation} \label{eq:resample}
    \mathbb{P}\left( \bbf =  \left(b^{(i)}_{1, (j,l)}, \dots, b^{(i)}_{q_{j,l}, (j,l)}  \right) \right) = w^{(i)},
\end{equation}
for all $i = 1, \dots, N$.
Note that the algorithm can be easily parallelized by drawing batches of samples on different cores.
This additional step would further reduce the computational times.

We explicit the BUIS algorithm on the simple hierarchy in Fig.~\ref{fig: simple tree}:
\begin{enumerate}
    \item Sample $(b_j^{(i)})_{i=1,\dots,N}$ from $\pi_{B_j}$, for $j=1,2,3,4$
    \item Compute the weights $(w^{(i)})_{i=1,\dots,N}$ with respect to $U_2$ as \[w^{(i)} = \pi_{U_2}\left( b_1^{(i)} + b_2^{(i)} \right)\]
    \item Sample $\left(\Bar{b}_1^{(i)}, \Bar{b}_2^{(i)}\right)_i$ with replacement from $\left( (b_1^{(i)}, b_2^{(i)}), w^{(i)} \right)_{i=1,\dots,N}$
    \item Repeat step $2$ and $3$ using $B_3, B_4$ and $U_3$ to get $\left(\Bar{b}_3^{(i)}, \Bar{b}_4^{(i)}\right)_i$
    \item 
    Set $\left(b_1^{(i)}, b_2^{(i)}, b_3^{(i)}, b_4^{(i)}\right)_i
    = \left(\Bar{b}_1^{(i)}, \Bar{b}_2^{(i)}, \Bar{b}_3^{(i)}, \Bar{b}_4^{(i)}\right)_i$
    and move to the next level 
    \item Compute the weights $(w^{(i)})_{i=1,\dots,N}$ with respect to $U_1$ as \[w^{(i)} = \pi_{U_1}\left( b_1^{(i)} + b_2^{(i)} + b_3^{(i)} + b_4^{(i)} \right)\]
    \item Sample $\left(\Bar{b}_1^{(i)}, \Bar{b}_2^{(i)}, \Bar{b}_3^{(i)}, \Bar{b}_4^{(i)} \right)_i$ with replacement from  $\left( (b_1^{(i)}, b_2^{(i)}, b_3^{(i)}, b_4^{(i)}), w^{(i)} \right)_i$
\end{enumerate}

In Appendix~\ref{sec: proof} we prove the following proposition:
\begin{proposition}\label{prop: alg}
The output of the BUIS algorithm is approximately a sample drawn from the reconciled distribution $\Tilde{\nu}$.
\end{proposition}

\subsection{Sample-based BUIS} 
\label{sec: buis w samples}

Sometimes the base forecasts are given as samples, without a parametric form;
this is the case of  models for time series of counts \citep{tscount} 
or based on deep learning \citep{salinas2020deepar}.
BUIS can reconcile also this type of base forecasts.
Since we only deal with one-dimensional densities to compute the weights, we use approximations based on samples.
For discrete distributions, we use the empirical distribution.
For continuous distributions, we use kernel density estimation \citep{chen2017tutorial}.
Therefore, we only need to replace line $4$ in Algorithm \ref{alg:BUIS}
with:\\

\begin{minipage}{10 cm}%
\centering
\begin{algorithmic}
\State \textbf{Sample} $\left( u_{j,l}^{(i)} \right)_{i=1,\dots,N}$ from $\hat{\pi}_{u_{j, l}}$
\State $\widecheck{\pi} \gets \textbf{Density Estimation} \left( \left( u_{j,l}^{(i)} \right)_{i=1,\dots,N} \right)$
\State $\widecheck{w}^{(i)} \gets \widecheck{\pi} \left(\sum_{t=1}^{q_{j,l}} b_{t, (j,l)}^{(i)}\right)$ \quad for  $i=1,\dots,N$\\
\end{algorithmic}
\end{minipage}

\noindent The sample-based algorithm becomes slightly slower due to the density estimation step. 

\subsection{More complex hierarchies: grouped time series}
\label{sec: grouped time series}

We refer to \textit{grouped} time series when the data structure does not disaggregate in a unique hierarchical manner \citep[Chapter~11]{hyndman2021forecasting}.
In this case, the aggregated series cannot be represented by a single tree, as a bottom node can have more than one parent.
For instance, consider a weekly time series, for which we compute the following temporal aggregates: $2$-weeks, $4$-weeks, $13$-weeks, $26$-weeks, $1$-year. A bottom node (weekly) is thus children of both the 
$2$-weeks and of the $13$-weeks aggregates.
This structure cannot be represented as a tree.

The BUIS algorithm, as described in Sec.~\ref{sec: buis algorithm}, requires that the hierarchy is a tree, so it cannot be used in this case.
Indeed, as highlighted in the proof, 
we need the independence of $\Bar{\bbf}_1, \dots, \Bar{\bbf}_{k_l}$ to multiply their densities. If the hierarchy is not a tree,
correlations between bottom variables are created when conditioning on the upper levels.

To overcome this problem, we proceed as follows. 
First, we find the largest sub-hierarchy within the group structure. 
For instance, in the example above, we consider the sub-hierarchy given by the bottom variables and by the $2$-weeks, $4$-weeks and $1$-year aggregates. 
All the other upper variables are then regarded as additional constraints.
We use the BUIS algorithm on the sub-hierarchy, obtaining a sample $\bbf$. 
Then, we compute the weights on $\bbf$ using the base distributions of the additional constraints. 
This is equivalent to performing a standard IS, where we use the output of BUIS on the hierarchical part as proposal distribution.
In this way, 
we reduce the dimension of the IS task from $n_u$, the total number of upper constraints, to the number of constraints that are not included in the sub-hierarchy:
in the above example, from $46$ to $6$.
We highlight that the distribution we sample from would be the same even with different choices of sub-hierarchies. 
However, picking the largest one is the best choice from a computational perspective.

\section{Experiments on synthetic data}
\label{sec: Experiments on synthetic data}

\begin{figure*}[!ht]
    \centering
    \includegraphics[width=0.8\textwidth]{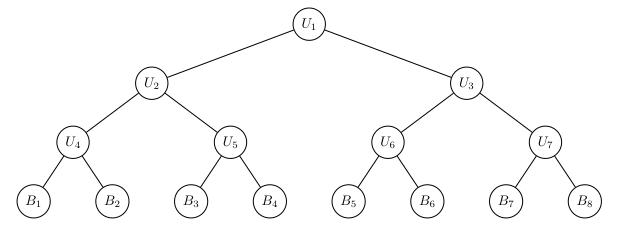}
    \caption{A binary hierarchy}
    \label{fig:bin hier}
\end{figure*}

We now empirically test the convergence of the BUIS algorithm to the true reconciled distribution. 
We compare BUIS with  IS and with the method
by \cite{corani2022probabilistic}, which we implement using the  library \textit{PyMC} \citep{salvatier2016probabilistic}. 
\textit{PyMC} adopts an adaptive Metropolis-Hastings algorithm \citep{haario2001adaptive} for discrete distributions and the No-U-Turn Sampler (NUTS, \cite{hoffman2014no}) 
for continuous distributions.
We performed experiments on the hierarchy of Fig.~\ref{fig:bin hier},
implementing the  IS and BUIS algorithms in Python. 

\subsection{Reconciling Gaussian forecasts}
\label{sec: exp synthetic gaussian}


\begin{table*}[t]
    \centering
    \begin{tabular}{c@{\hskip 1cm}c@{\hskip 1cm}ccc}
        \toprule
        {} & {} & \multicolumn{3}{c}{Number of samples}\\
        \cmidrule{3-5}\\
        \addlinespace[-10pt]
        & & $10^4$ & $10^5$ & $10^6$\\
        \midrule
        \multirow{2}{*}{\textit{Gaussian}} & \textbf{IS} & 0.01 {\footnotesize $\pm$ 0.00} & 0.06 {\footnotesize $\pm$ 0.02} & 0.66 {\footnotesize $\pm$ 0.17} \\
                                           & \textbf{BUIS} & 0.02 {\footnotesize $\pm$ 0.00} & 0.15 {\footnotesize $\pm$ 0.00} & 2.20 {\footnotesize $\pm$ 0.02} \\
        \midrule
        \multirow{3}{*}{\textit{Poisson}} & \textbf{IS} & 0.01 {\footnotesize $\pm$ 0.01} & 0.09 {\footnotesize $\pm$ 0.01} & 1.01 {\footnotesize $\pm$ 0.03} \\
        & \textbf{BUIS}    & 0.02 {\footnotesize $\pm$ 0.01} & 0.19 {\footnotesize $\pm$ 0.03} & 2.68 {\footnotesize $\pm$ 0.41} \\[3pt]
        & \makecell{\textbf{sample-based} \\ \textbf{BUIS}} & 0.03 {\footnotesize $\pm$ 0.00} & 0.24 {\footnotesize $\pm$ 0.01} & 3.49 {\footnotesize $\pm$ 0.09} \\
        \bottomrule
    \end{tabular}
    \caption{\label{tab: times}Average computational times with the standard deviations (in seconds). The average times for \textit{PyMC} ($4$ chains with $5,000$ samples each) are: $26.81$ {\footnotesize $\pm$ 2.38} (Gaussian), 26.26 {\footnotesize $\pm$ 4.14} (Poisson).}
\end{table*}

Dealing with Gaussian base forecasts, the 
reconciled distribution can be obtained in closed form \citep{corani_ecml_reconc}. 
We can thus check how the various algorithms approximates the exact
solution.
We set on each bottom node a Gaussian distribution with mean randomly chosen in the interval $[5,10]$, and standard deviation $\sigma_b = 2$. We denote by $\bm{\mu}_b \in \rr_+^8$ the vector of the base bottom means.
We induce incoherence by setting the means of the base forecast of the upper variables as  $\bm{\mu}_u = (1+\epsilon) \Abf \bm{\mu}_b$, where $\Abf$ is the aggregating matrix and $\epsilon$ is the  incoherence level; 
we consider $\epsilon \in \{ 0.1, \; 0.3, \: 0.5 \}$.
Hence, if $\epsilon$ =0.3  the base upper means are $30\%$ greater than the sum of the corresponding base bottom means.
We set $\sigma_u = 3$ as  standard deviation for the base forecast of  each upper variable.

\begin{figure}[!h]
     \centering
     \begin{subfigure}[b]{0.48\textwidth}
         \centering
         \includegraphics[width=\textwidth]{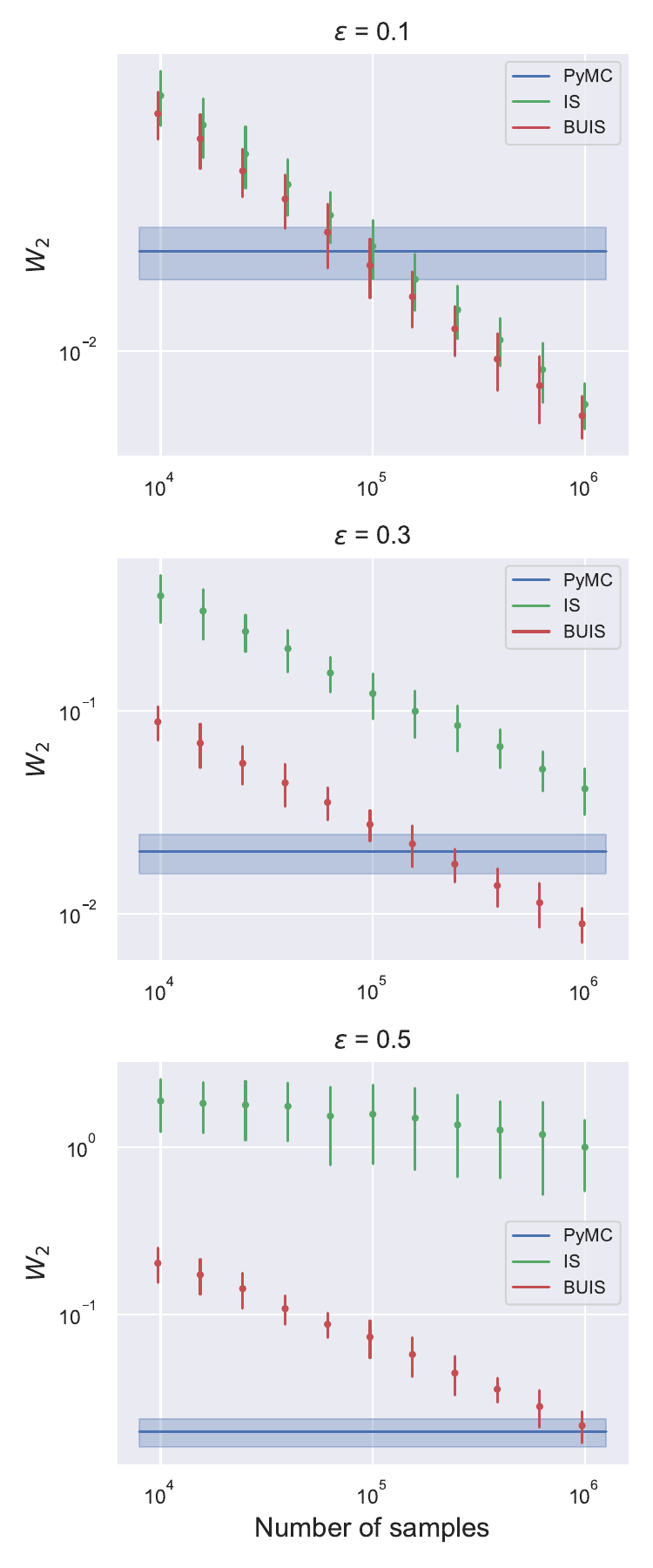}
         \caption{Gaussian distributions}
         \label{fig: W2 gaussian}
     \end{subfigure}
     \hfill
     \begin{subfigure}[b]{0.48\textwidth}
         \centering
         \includegraphics[width=\textwidth]{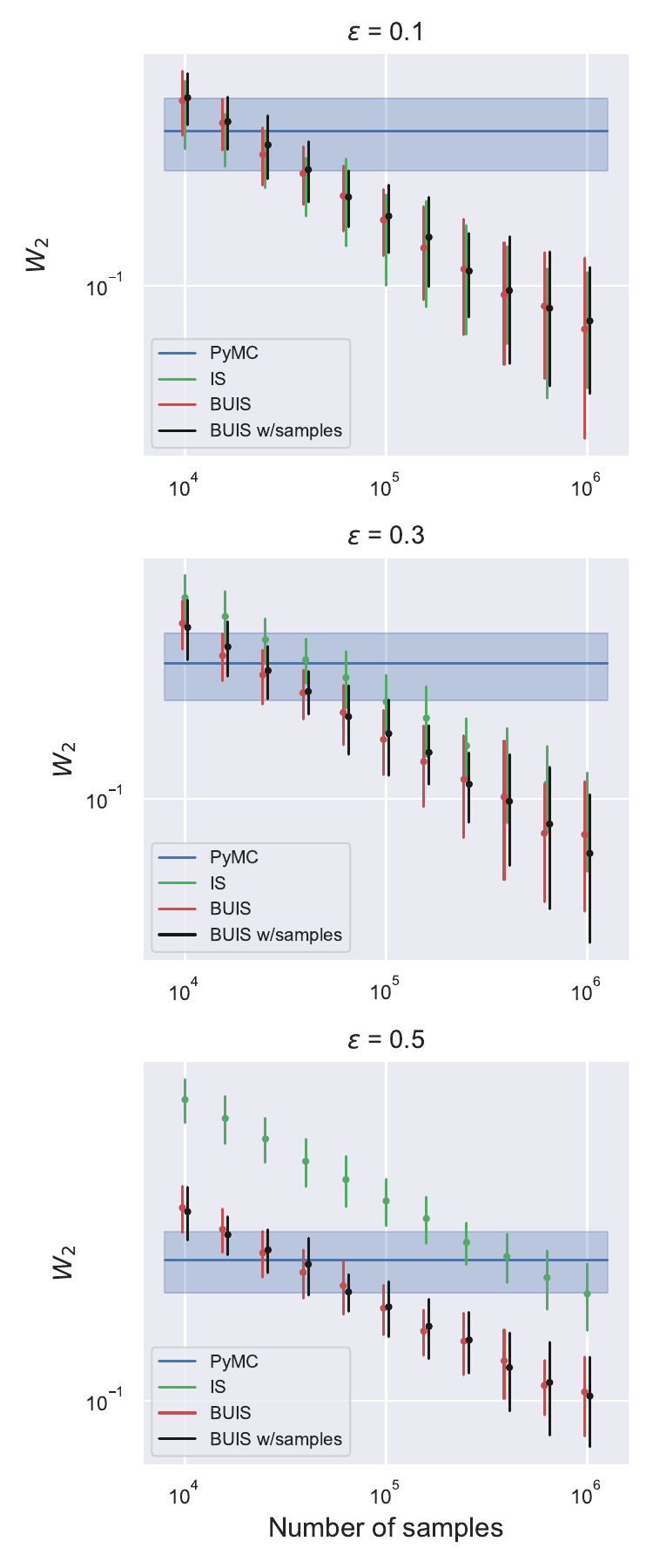}
         \caption{Poisson distributions}
         \label{fig: W2 poisson}
     \end{subfigure}
        \caption{Wasserstein distance between true and empirical distributions. The axes are logarithmic.}
        \label{fig: W2}
\end{figure}

We run \textit{PyMC} with $4$ chains with $5,000$ samples each.
For IS and BUIS, we run multiple experiments, drawing each time a different number of samples, ranging from $10^4$ to $10^6$.
We repeat each experiment $30$ times.
We then compute the $2$-Wasserstein distance \citep{panaretos2019statistical} between the true reconciled distribution, obtained analytically, and the empirical distributions obtained via sampling.
The results are reported in Fig.~\ref{fig: W2 gaussian}, where we also show the $95\%$ confidence interval over the 30 experiments. 
Note that the axes are in logarithmic scale.

As expected, the performance of IS and BUIS depends on the incoherence level $\epsilon$.
This behavior also affects BUIS, which is based on importance sampling.
However,  BUIS is significantly more robust than IS, and it works effectively even with extreme  incoherence level  such as $\epsilon = 0.5$.
As the number of samples grows, the performance of BUIS improves, eventually outperforming the reference method based on \textit{PyMC}.
We confirm the results by computing the percentage error on the reconciled mean (Appendix~\ref{sec: app percentage error}).
Even with an extreme incoherence level, $\epsilon = 0.5$, the percentage error on the mean obtained with $10^6$ samples from BUIS is negligible ($<0.1\%$) and comparable to \textit{PyMC}. In the same setup IS achieves an error greater than $5\%$. 

Both IS and BUIS are substantially faster than \textit{PyMC} (Table \ref{tab: times}). 
The computational time of BUIS with $10^5$ samples is two orders of magnitude smaller than \textit{PyMC}, while achieving comparable performances. 
Note that here BUIS is running on a single core.
An insight about the reasons of such a speedup is given in Appendix~\ref{sec: appendix pymc3/is comparison}, where we provide a detailed comparison between IS and a bare-bones implementation of MCMC on a simple hierarchy.

We also conduct similar experiments using a larger hierarchy; the results, reported in Appendix~\ref{sec: app large hier}, confirm that the BUIS is robust and computationally efficient.

\subsection{Reconciling Poisson forecasts}

We now consider discrete base forecasts.
We set a Poisson distribution on each bottom variable, with mean randomly chosen in the interval $[5,10]$. We denote by $\bm{\lambda}_b \in \rr_+^8$ the vector of the base bottom means. 
As before, for each incoherence level $\epsilon \in \{0.1, \; 0.3, \; 0.5\}$, we set the mean of the upper variables as  $\bm{\lambda}_u = (1+\epsilon) \Abf \bm{\lambda}_b$. 
In the Poisson case, the reconciled distribution cannot be analytically computed. 
We thus run an extensive experiment using \textit{PyMC}, with $20$ chains with $50,000$ samples each. 
We consider these samples as the true reconciled distribution.

We run the same experiments described in Sec.~\ref{sec: exp synthetic gaussian}.
Since probabilistic forecasts of count time series are typically given as  samples \citep{tscount}, 
we also run sample-based BUIS (Sec.~\ref{sec: buis w samples}): we assume that the parametric form of the base distribution is unknown, and that only samples are available. 

The $2$-Wasserstein distances are reported in Fig.~\ref{fig: W2 poisson}.
As the number of samples grows, BUIS and sample-based BUIS eventually outperform \textit{PyMC}, for all levels of incoherence.
As for the Gaussian case, the performance of IS deteriorates for larger values of the incoherence level. 
The results are confirmed by the percentage error on the reconciled mean (Appendix~\ref{sec: app percentage error}), which is lower than $0.2\%$ for BUIS with $10^5$ samples and about $0.4\%$ for \textit{PyMC}.

In Table \ref{tab: times} we show the average computational times.
Sample-based BUIS is slightly slower than BUIS because of the density estimation step.
Note that, using $10^5$ samples, BUIS and sample-based BUIS are 2 orders of magnitude faster than \textit{PyMC}, while achieving a better performance for all incoherence levels.


\section{Experiments on real data}
\label{sec: Experiments}

\begin{table*}[!h]
    \centering
\begin{tabular}{ccccc}
\toprule
   &   & \textit{N} vs \textit{base} & \textit{NB} vs \textit{base} & \textit{samples} vs \textit{base} \\
\textbf{metric} & hier-level &           &            &                 \\
\midrule
\rowcolor{lightblue} \textbf{ES} &  &  0.07 &   0.52 &  \textbf{0.53} \\
\midrule
\textbf{MASE} & Monthly &     -1.02 &   \textbf{0.14} &           0.13 \\
   & 2-Monthly &     -0.53 &       0.25 &       \textbf{0.27} \\
   & Quarterly &     -0.42 &       0.21 &           \textbf{0.26} \\
   & 4-Monthly &     -0.40 &       0.16 &       \textbf{ 0.21} \\
   & Biannual &     -0.33 &       0.14 &          \textbf{ 0.16} \\
   & Annual &     -0.26 &     \textbf{  0.18} &            0.17 \\
   \rowcolor{lightblue} & \textit{average} &     -0.49 &       0.18 &         \textbf{   0.20} \\
  \midrule
\textbf{MIS} & Monthly &     -0.08 &       0.45 &       \textbf{ 0.63} \\
   & 2-Monthly &      0.28 &       0.45 &      \textbf{ 0.56} \\
   & Quarterly &      0.22 &       0.43 &       \textbf{ 0.46} \\
   & 4-Monthly &      0.03 &       0.35 &      \textbf{0.36} \\
   & Biannual &     -0.07 &    \textbf{0.37} &            0.26 \\
   & Annual &     -0.17 &     \textbf{0.40} &            0.22 \\
   \rowcolor{lightblue} & \textit{average} &      0.03 &       0.41 &           \textbf{ 0.42} \\
\bottomrule
\end{tabular}
    \caption{\label{tab:Carparts skill scores vs base} Skill scores on the time series extracted from \textit{carparts}, detailed by each level of the hierarchy}
\end{table*}

\begin{table*}[!h]
    \centering
\begin{tabular}{ccccc}
\toprule
   &   & N vs base & NB vs base & samples vs base \\
\textbf{metric} & \textbf{hier-level} &           &            &                 \\
\midrule
\rowcolor{lightblue} \textbf{ES} &  &   0.08 &   0.11 &    \textbf{0.15} \\
\midrule
\textbf{MASE} & Weekly &     -0.63 &    \textbf{0.14} &      \textbf{0.14} \\
   & 2-Weekly &     -0.40 &   \textbf{0.16} &            0.14 \\
   & 4-Weekly &     -0.22 &   \textbf{0.13} &            0.12 \\
   & Quarterly &     -0.10 &       0.01 &       \textbf{0.04} \\
   & Biannual &      0.01 &       0.07 &      \textbf{0.15} \\
   & Annual &     -0.05 &      -0.00 &      \textbf{0.04} \\
   \rowcolor{lightblue} & \textit{average} &     -0.23 &       0.08 &            \textbf{0.10} \\
\midrule
\textbf{MIS} & Weekly &     -0.06 &   \textbf{0.46} &            0.45 \\
   & 2-Weekly &      0.08 &       0.33 &      \textbf{0.34} \\
   & 4-Weekly &      0.03 &       0.19 &     \textbf{0.25} \\
   & Quarterly &     -0.15 &      -0.11 &    \textbf{-0.08} \\
   & Biannual &     -0.34 &      -0.27 &     \textbf{-0.21} \\
   & Annual &     -0.33 &      -0.23 &     \textbf{-0.22} \\
   \rowcolor{lightblue} & \textit{average} &     -0.13 &       0.06 &           \textbf{ 0.09} \\
\bottomrule
\end{tabular}
\caption{\label{tab:Syph skill scores vs base} Skill scores on the time series extracted from \textit{syph}, detailed by each level of the hierarchy}
\end{table*}

We now perform probabilistic reconciliation on temporal hierarchies, using time series extracted from two different data sets: \textit{carparts}, available from the R package \textit{expsmooth} \citep{expsmooth}, and \textit{syph}, available from the R package \textit{ZIM} \citep{zim_manual}.

The \textit{carparts} data set is about monthly sales of car parts. 
As in \cite[Chapter~16]{hyndman2008forecasting}, we remove time series with missing values, with less then $10$ positive monthly demands and with no positive demand in the first $15$ and final $15$ months. 
After this selection, there are $1046$ time series left.
Note that we use less restrictive criteria in the selection of the time series than \cite{corani2022probabilistic}, where only $219$ time series from \textit{carparts} were considered.
Monthly data are aggregated into $2$-months, $3$-months, $4$-months, $6$-months and $1$-year levels. 

The \textit{syph} data set is about the weekly number of syphilis cases in the United States. 
We remove the time series with ADI greater than $20$.
The ADI is computed as $\ADI = \frac{\sum_{i=1}^P p_i}{P}$, where $p_i$ is the time period between two non-zeros values and $P$ is the total number of periods \citep{SyntetosBoylan2005}.
We also remove the time series corresponding to the total number of cases in the US. 
After this selection, there are $50$ time series left.
Weekly data are aggregated into $2$-weeks, $4$-weeks, $13$-weeks, $26$-weeks and $1$-year levels. 

For both data sets, we fit a generalized linear model with the \textit{tscount} package \citep{tscount}. 
We use a negative binomial predictive distribution, with a first-order regression on past observations.
The test set has length $1$ year for both data sets. 
We thus compute up to $12$ steps ahead at monthly level, and up to $52$ steps ahead at weekly level.
Probabilistic forecasts are returned in the form of samples. 

Reconciliation is performed in three different ways.
In the first case, we fit a Gaussian distribution on the returned samples. Then, we follow \citep{corani_ecml_reconc} to analytically compute the Gaussian reconciled distribution.
In the second case, we fit a negative binomial distribution on the samples, and we reconcile using the BUIS algorithm.
Since these  are grouped time series rather than hierarchical time series,
we use the method of Sec.~\ref{sec: grouped time series} for grouped time series.
Finally, we use the sample-based BUIS (Sec.~\ref{sec: buis w samples}), without fitting a parametric distribution.
Although the sample-based algorithm is slightly slower, this method yields a computational gain over BUIS, as fitting a negative binomial distribution on the samples requires about $1.2$ s for the monthly hierarchy and $3.9$ s for the weekly hierarchy.
We refer to these methods, respectively, as \textit{N}, \textit{NB}, and \textit{samples}.
Furthermore, we denote by \textit{base} the unreconciled forecasts.

We use different indicators to assess the performance of each method.
The mean scaled absolute error (MASE) \citep{hyndman2006another} is defined as
\[
\text{MASE} = \frac{\text{MAE}}{Q},
\]
where 
$\text{MAE} = \frac{1}{h} \sum_{j=1}^h \lvert y_{t+j} - \hat{y}_{t+j\mid t}\rvert$ and $Q =  \frac{1}{T-1} \sum_{t=2}^T \lvert y_t - y_{t-1}\rvert$.
Here, $y_t$ denotes the value of the time series at time $t$, while $\hat{y}_{t+j\mid t}$ denotes the point forecast computed at time $t$ for time $t+j$.
The median of the distribution is used as point forecast, since it minimizes MASE \citep{kolassa2016evaluating}.

The mean interval score (MIS) \citep{gneiting2011quantiles} is defined, for any $\alpha \in (0,1)$, as
\[
\text{MIS} = (u - l) + \frac{2}{\alpha} (l - y) \mathbb{1}(y<l) + \frac{2}{\alpha} (y-u) \mathbb{1}(y>u),
\]
where $l$ and $u$ are the lower and upper bounds of the $(1-\alpha)$ forecast coverage interval and $y$ is the actual value of the time series. In the following, we use $\alpha = 0.1$.
MIS penalizes wide prediction intervals, as well as intervals that do not contain the true value. 


Finally, the Energy score \citep{szekely2013energy} is defined as
\[
ES(P,\ybf) = \E_P\left[\|\ybf-\sbf\|^{\alpha}\right] 
- \frac{1}{2} \E_P\left[\|\sbf-\sbf'\|^{\alpha}\right],
\]
where $P$ is the forecast distribution on the whole hierarchy, $\sbf, \sbf' \sim P$ are a pair of independent random variables and $\ybf$ is the vector of the actual values of all the time series.
The energy score is a proper scoring rule for distributions defined on the entire hierarchy \citep{panagiotelis2022probabilistic}.
We compute $ES$, with $\alpha = 2$, using samples, as explained in \cite{wickramasuriya2021probabilistic}.

We use the skill score to compare the performance of a method with respect to a baseline method, in terms of percentage improvement.
We use \textit{base} as baseline method.
For example, the skill score of \textit{NB} on MASE is given by
\[
\text{Skill}(\textit{NB}, \textit{base}) 
= \frac{\text{MASE}(\textit{base}) - \text{MASE}(\textit{NB})}
{\left(\text{MASE}(\textit{base}) + \text{MASE}(\textit{NB})\right) / 2}.
\]
Note that the skill score is symmetric and scale-independent. 
For each level, we compute the skill score for each forecasting horizon, and take the average.

The skill scores for \textit{carparts} are reported in Table \ref{tab:Carparts skill scores vs base}.
Both \textit{NB} and \textit{samples} methods yield a significant improvement for all the indicators, and for all the hierarchy levels. 
For both methods, the average improvement is about $20\%$ for MASE, $40\%$ for MIS and $50\%$ for ES.
The skill scores for \textit{syph} are reported in Table \ref{tab:Syph skill scores vs base}.
As before, the average improvement of \textit{NB} and \textit{samples} is significant for all indicators.
For both datasets, the \textit{N} method performs poorly, in many cases yielding negative skill scores.
As observed in \cite{corani2022probabilistic}, this method does not capture the asymmetry of the base forecasts.
Finally, \textit{samples} appears to perform better that \textit{NB}.
Indeed, the step of fitting a Negative Binomial distribution on the forecast samples may yield an additional source of error.

\section{Conclusions}
\label{sec: conclusion}


Our approach to probabilistic reconciliation based on conditioning allows to treat continuous and discrete forecast distributions in a unified framework.
Moreover, the proposed BUIS is able to efficiently sample from 
continuous and discrete predictive distributions, provided in parametric form or as samples. 
We make available the BUIS algorithm within  the \textit{R} package \textit{bayesRecon} \citep{bayesRecon}.

A future research direction is how to relax the assumption of conditional independence of the base forecasts.
A second one is to study the implications of ignoring the behavior of the base forecast distribution outside the coherent subspace,
which is a feature of reconciliation via conditioning and constitutes a major difference from reconciliation via projection.

\section{Acknowledgements}
Work partially funded by the Swiss National Science Foundation (grant 200021\_212164/1) and by the Hasler foundation (project 23057).


\def\bibfont{\footnotesize}
\bibliographystyle{abbrvnat}
\bibliography{biblio}


\newpage

\appendix

\section{Proofs}
\label{sec: appendix proofs}

\begin{proposition}\label{prop: pushforward bij}
Let $s: X \to Y$ be a measurable bijection between two measure spaces $(X,\mathcal{X})$ and $(Y,\mathcal{Y})$. 
Then, the pushforward $s_{\#}: \pp(X) \to \pp(Y)$ is a bijection, with inverse given by $(s^{-1})_{\#}$.
\end{proposition}

\begin{proof}
First, we recall that the pushforward $s_{\#}$ is defined, for any $\nu \in \pp(X)$ and $F \in \mathcal{Y}$, as
\[
s_{\#}\nu(F) = \nu(s^{-1}(F)).
\]
Hence, for any $\nu \in \pp(X)$ and $G \in \mathcal{X}$, we have
\begin{align*}
\big((s^{-1})_{\#} \circ s_{\#}\big)\nu \, (G)
&= (s^{-1})_{\#} \big( s_{\#}\nu \big) \, (G) \\
&= s_{\#}(\nu) \big( (s^{-1})^{-1}(G) \big)  \\
&= s_{\#}(\nu) \big( s(G) \big) \\
&= \nu\big(s^{-1}(s(G))\big) \\
&= \nu(G),
\end{align*}
and therefore $(s^{-1})_{\#} \circ s_{\#}$ is the identity map.  
Analogously, for any $\mu \in \pp(Y)$ and $F \in \mathcal{X}$, we have
\begin{align*}
\big(s_{\#} \circ (s^{-1})_{\#}\big)\mu \, (F)
&= s_{\#} \big((s^{-1})_{\#}\mu \big) \, (F) \\
&= (s^{-1})_{\#}(\mu) \big( s^{-1}(F) \big)  \\
&= \mu\big((s^{-1})^{-1}\big(s^{-1}(F)\big)\big) \\
&= \mu(s(s^{-1}(F))) \\
&= \mu(F).
\end{align*}
\end{proof}

\bigskip

\begin{proposition}\label{prop: change of variable}
Let $\hat{\pi}$ be the joint density of the random vector $(\UH,\BH)$.
Then, the density of $(\Zbf,\BH)$, where $\Zbf:= \UH - \Abf \BH$, is given by
\[
\pi_{(\Zbf,\BH)}(\zbf,\bbf) = \hat{\pi}(\zbf + \Abf \bbf, \bbf).
\]
\end{proposition}

\begin{proof}
The joint density of $(\Zbf,\BH)$ can be computed using the rule of change of variables 
\cite[Chapter~17]{billingsley2008probability}.
Let $\Hbf:\rr^n \to \rr^n$ be defined as
\[
\Hbf: \begin{bmatrix} \ubf \\ \bbf \end{bmatrix} 
\to \begin{bmatrix} \ubf - \Abf \bbf \\ \bbf \end{bmatrix}.
\]
$\Hbf$ is invertible, with inverse given by
\[
\Hbf^{-1}: \begin{bmatrix} \zbf \\ \bbf \end{bmatrix} 
\to \begin{bmatrix} \zbf + \Abf \bbf \\ \bbf \end{bmatrix},
\]
and we have that
\[
\begin{vmatrix} J \Hbf^{-1} (\bbf,\zbf) \end{vmatrix}
= \begin{vmatrix} \textbf{I} & \Abf^T \\ \textbf{0} & \textbf{I} \end{vmatrix}
= 1.
\]
Then, the joint density of $(\Zbf,\BH)$ is given by
\begin{align*}
\pi_{(\Zbf,\BH)}(\zbf,\bbf)
&= \hat{\pi}\big(\Hbf^{-1}(\zbf,\bbf)\big) \cdot \begin{vmatrix} J \Hbf^{-1} (\zbf,\bbf) \end{vmatrix} \\
&= \hat{\pi}(\zbf + \Abf \bbf, \bbf).
\end{align*}
\end{proof}

\section{Proof of BUIS algorithm}
\label{sec: proof}

We show that the output $\big(\bbf^{(i)}\big)_i$ of the BUIS algorithm is approximately a sample drawn from the target distribution $\nutil$.

From \eqref{eq: mutil indep}, and from Assumption \ref{assumptions: independent}, we have that 
\begin{align*}
\Tilde{\pi}(\bbf) &\propto \hat{\pi}_b(\bbf)  \cdot \hat{\pi}_u(\Abf \bbf)  \\
&= \prod_{t=1}^{n_b} \pi_{b_t}(b_t) \cdot \prod_{l=1}^L \prod_{j=1}^{k_l} \pi_{u_{j,l}} \bigg( \sum_{k=1}^{q_{j,l}} b_{k, (j, l)} \bigg),
\end{align*}
where we are using the notation of Sec.~\ref{sec: buis algorithm}.
The initial distribution of the sample
$\big(\bbf^{(i)}\big)_{i=1,\dots,N}$ is given by $\hat{\pi}_b = \prod_{t=1}^{n_b} \pi_{b_t}(b_t)$. 
We show that each iteration of the algorithm corresponds to multiplying by a $\pi_{u_{j,l}} \bigg( \sum_{k=1}^{q_{j,l}} b_{k, (j, l)} \bigg)$ term.

Let $\pi_X$ be a density over $\rr^d$, and $w: \rr^d \to \rr$ a continuous function. 
Let $X_1,\dots,X_N$ be independent samples from $\pi_X$, and compute the unnormalized weights $(\hat{w}^{(i)})_{i=1,\dots,N}$ as $\hat{w}^{(i)} = w(X_i)$. 
Then, if we draw $Y_1,\dots,Y_m$ from the discrete distribution given by
\[
\mathbb{P}\left( Y =  X_i \right) = w^{(i)}, \qquad i=1,\dots,N, 
\]
where $w^{(i)} = \frac{\hat{w}^{(i)}}{\sum_{j=1}^N \hat{w}^{(j)}}$,
then $(Y_i)_{i=1,\dots,m}$ is approximately an IID sample from the density $\pi_Y(x) \propto \pi_X(x) \cdot w(x)$. 
This technique is known as importance resampling or weighted bootstrap \citep{smith1992bayesian}. 
The same holds also for discrete distributions, using the pmf instead of the density.

Hence, 
if we compute the weights $w^{(i)}$'s as in the algorithm and sample $\big(\Tilde{\bbf}_j^{(i)}\big)_i$ from \eqref{eq:resample}, 
it is approximately equivalent to sampling from $\hat{\pi}_b(\bbf) \cdot \pi_{u_{j, l}}\left(\sum_{t=1}^{q_{j,l}} b_t\right)$,
where $\hat{\pi}_b$ is the original density of $\left(b_{1, (j,l)}, \dots, b_{q_{j,l}, (j,l)}\right)$.
In other words, the weighting-resampling step corresponds to multiplying the density of the sample by a $\pi_{u_{j, l}}\left(\sum_{t=1}^{q_{j,l}} b_t\right)$ term.

Finally, note that in this way we are conditioning with respect to $u_{j, l}$. After the weighting-resampling step, $\left(b_{1, (j,l)}, \dots, b_{q_{j,l}, (j,l)}\right)$ are correlated.
Since 
the hierarchy is given by a tree, we are guaranteed that 
for any level $l$ and for all $j=1,\dots,k_l$,
$\Tilde{\bbf}_j$ only depends on 
$b_{1, (j,l)}, \dots, b_{q_{j,l}, (j,l)}$, $u_{j, l}$ and each upper variable that is under $u_{j, l}$. 
From Assumption~\ref{assumptions: independent}, we have that $\Tilde{\bbf}_1, \dots, \Tilde{\bbf}_{k_l}$ are independent.
Hence, the density of 
$\left[\Tilde{\bbf}_1,\dots,\Tilde{\bbf}_{k_l}\right]$ is given by the product of the densities of all $\Tilde{\bbf}_j$'s, and the proof is concluded.


\section{MCMC-IS comparison}
\label{sec: appendix pymc3/is comparison}

In order to fully understand the reasons for the significant difference in computational time between the MCMC and the IS approach, 
we compare the two methods on a minimal example.
Le us consider a hierarchy given by two bottom variables, $b_1$ and $b_2$, and just one upper variable $u$, which is the sum of $b_1$ and $b_2$. 
We set a Gaussian distribution for each variables. 

We implement a simple Metropolis-Hastings algorithm with a Gaussian proposal distribution with fixed variance $\tau I$ to sample from the reconciled distribution $\Tilde{\pi}(\bbf) = \pi_{b_1}(b_1) \cdot \pi_{b_2}(b_2) \cdot \pi_{u}(b_1 + b_2)$. 
The algorithm reads as follows:\\

\begin{algorithmic} 
\State \textbf{Initialize} $\bbf^{(0)}$
\For {$j=1,\dots,N$}
    \State \textbf{Sample} $\ybf^{(j)} \sim \mathcal{N}(\bbf^{(j-1)}, \tau I)$
    \State $\alpha \gets \min\left(1, \frac{\Tilde{\pi}(\ybf^{(j)})}{\Tilde{\pi}\left(\bbf^{(j-1)}\right)}\right)$
    \State $u \gets \Unif(0,1)$
    \If {$u < \alpha$}
        \State $\bbf^{(j)} \gets \ybf^{(j)}$
    \Else
        \State $\bbf^{(j)} \gets \bbf^{(j-1)}$
    \EndIf
\EndFor
\State \Return $\big(\bbf^{(i)}\big)_i$
\end{algorithmic}

\vspace{5mm}

On a standard laptop, it takes about 4 seconds to get $10,000$ samples from $\Tilde{\pi}$. In particular, most of the time is employed by the computation of the acceptance probability $\alpha$, which requires about $3.7 \cdot 10^{-4}$ seconds per loop. Sampling from the proposal distribution only requires about $3 \cdot 10^{-5}$ seconds.

We then implement an IS algorithm on the same hierarchy, using Python:\\

\begin{algorithmic} 
\State \textbf{Sample} $\bbf^{(1)},\dots,\bbf^{(N)} \overset{\text{IID}}{\sim} \hat{\pi}_b$
\State $w_i \gets \hat{\pi}_u\left(b_1^{(i)}+b_2^{(i)}\right)$
\State \Return $\big(\bbf^{(i)}, w_i\big)_i$
\end{algorithmic}

\bigskip

It takes about $7 \cdot 10^{-3}$ seconds to draw $100,000$ IID samples from $\hat{\pi}_b$, and about the same time to compute all the weights.
The significant improvement in computational time is due to the fact that both sampling and computation of the weights are done simultaneously for all the samples, rather than sequentially as in MCMC.


\section{Additional results on synthetic data}
\label{sec: appendix extended results}

\subsection{Percentage error on the mean}
\label{sec: app percentage error}

Besides computing the 2-Wasserstein distance between the true reconciled distribution and the empirical reconciled distribution obtained via sampling (Sect.~\ref{sec: Experiments on synthetic data}),
we also compute the error on the reconciled mean.  
More precisely, if we denote by $m_i$ the true mean and by $\overline{m}_i$ the sample mean, we compute the average percentage error as:
\begin{equation*}
    \frac{1}{n} 
    \sum_{i=1}^n \frac{\lvert m_i - \overline{m}_i \rvert}{m_i} \cdot 100,
\end{equation*}
where $n$ is the number of nodes of the hierarchy.
The average percentage errors are reported in Table~\ref{tab: mape}.

\begin{table*}[h!]
    \centering
    \begin{tabular}{c@{\hskip 1cm}c@{\hskip 1cm}cccc}
        \toprule
        & & & \multicolumn{3}{c}{$\epsilon$}\\
        \cmidrule{4-6}\\
        \addlinespace[-10pt]
        & & & $0.1$ & $0.3$ & $0.5$\\
        \midrule
        \multirow{7}{*}{\textit{Gaussian}} &
        \multirow{3}{*}{\textbf{IS}} & $10^4$ samples & 0.39\% & 2.02\% & 11.3\% \\
                                  & & $10^5$ samples & 0.13\% & 0.6\% & 8.96\% \\
                                  & & $10^6$ samples & 0.04\% & 0.22\% & 5.69\% \\[5pt]
        & \multirow{3}{*}{\textbf{BUIS}} & $10^4$ samples & 0.34\% & 0.45\% & 0.92\% \\
                                   &  & $10^5$ samples & 0.12\% & 0.14\% & 0.34\% \\
                                   &  & $10^6$ samples & 0.04\% & 0.05\% & 0.09\% \\[5pt]
        & \textbf{\textit{PyMC}} & $4 \times 5000$ samples & 0.11\% & 0.09\% & 0.07\% \\
        \midrule
        \multirow{7}{*}{\textit{Poisson}} &
        \multirow{3}{*}{\textbf{IS}} & $10^4$ samples & 0.44\% & 0.75\% & 2.24\% \\
                                    & & $10^5$ samples & 0.13\% & 0.25\% & 0.65\% \\
                                   &  & $10^6$ samples & 0.06\% & 0.09\% & 0.23\% \\[5pt]
        & \multirow{3}{*}{\textbf{BUIS}} & $10^4$ samples & 0.5\% & 0.58\% & 0.67\% \\
                                    & & $10^5$ samples & 0.16\% & 0.16\% & 0.21\% \\
                                   & & $10^6$ samples & 0.06\% & 0.07\% & 0.09\% \\[5pt]
        & \multirow{3}{*}{\makecell{\textbf{sample-based} \\ \textbf{BUIS}}} & $10^4$ samples & 0.52\% & 0.55\% & 0.59\% \\
                                    & & $10^5$ samples & 0.17\% & 0.17\% & 0.21\% \\
                                   & & $10^6$ samples & 0.07\% & 0.07\% & 0.08\% \\[5pt]
        & \textbf{\textit{PyMC}} & $4 \times 5000$ samples & 0.09\% & 0.07\% & 0.07\% \\
        \bottomrule
    \end{tabular}
    \caption{\label{tab: mape}Average percentage errors on the reconciled mean.}
\end{table*}

\subsection{Large hierarchy}
\label{sec: app large hier}

\begin{figure*}[h]
    \centering
    \includegraphics[width=\textwidth]{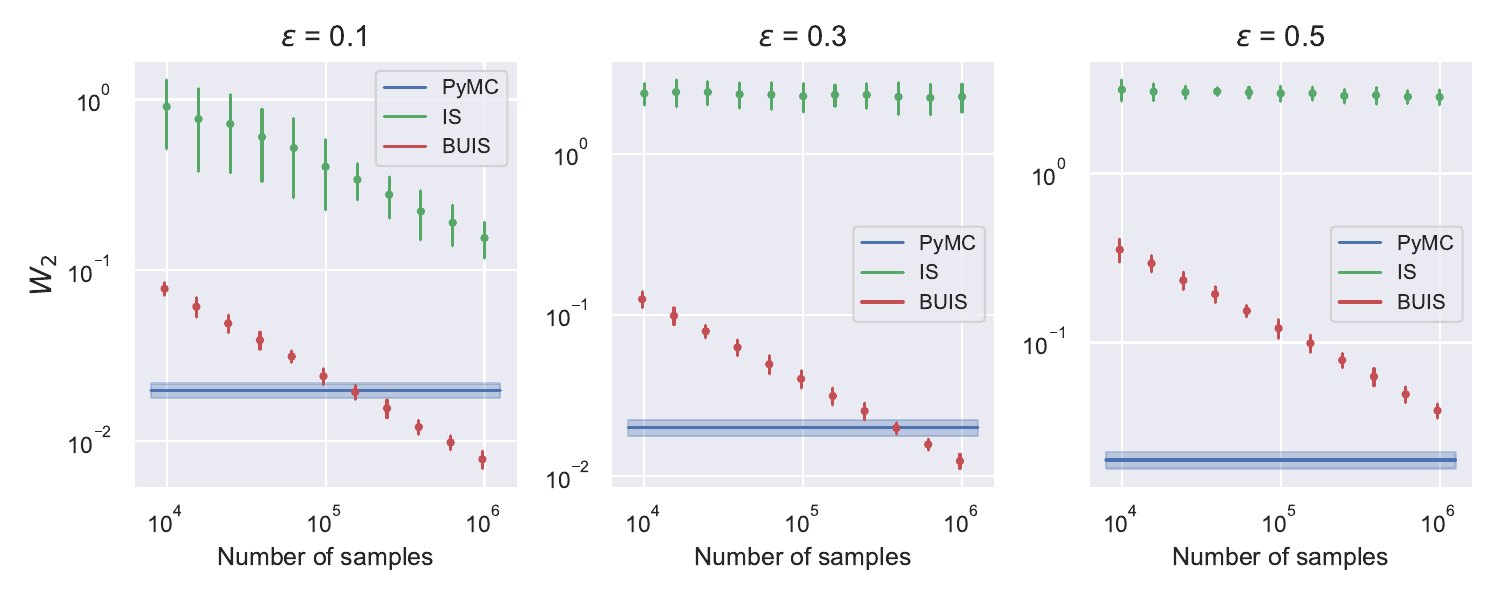}
    \caption{Wasserstein distance between true and empirical distributions (Gaussian case, large hierarchy). The axes are logarithmic.}
    \label{fig: W2 gaussian large}
\end{figure*}

We test the IS, BUIS, and \textit{PyMC} algorithms on a larger hierarchy.
We set a binary hierarchy, similar to that of Fig.~\ref{fig:bin hier}, but with $5$ levels:
hence, there are $32$ bottom and $31$ upper nodes.
We use the same procedure described in Sect.~\ref{sec: exp synthetic gaussian} to set the Gaussian base forecasts.
Using BUIS with $10^5$ samples we achieve a small average percentage error ($< 0.5\%$) on the reconciled means (Table~\ref{tab: mape large hier}), even with a large incoherence ($\epsilon = 0.5$).
On the other hand, the error using IS is over $20\%$, even with $10^6$ samples, proving that IS is not able to scale to large hierarchies.
The results are confirmed by the plot of the 2-Wasserstein distance (Fig.~\ref{fig: W2 gaussian large}). 
In conclusion, BUIS is able to correctly sample from the reconciled distribution, even in case of rather big hierarchies ($\sim 60$ nodes) and large incoherence level ($\epsilon = 0.5$), while providing an impressive gain in terms of computational time with respect to \textit{PyMC} (Table~\ref{tab: times large hier}).

\begin{table*}[h]
    \centering
    \begin{tabular}{c@{\hskip 1cm}cccc}
        \toprule
        & & \multicolumn{3}{c}{$\epsilon$}\\
        \cmidrule{3-5}\\
        \addlinespace[-10pt]
        & & $0.1$ & $0.3$ & $0.5$\\
        \midrule
        \multirow{3}{*}{\textbf{IS}} & $10^4$ samples & 6.72\% & 17.13\% & 23.96\% \\
                                  & $10^5$ samples & 2.78\% & 16.05\% & 21.61\% \\
                                  & $10^6$ samples & 1.0\% & 16.25\% & 20.11\% \\[5pt]
        \multirow{3}{*}{\textbf{BUIS}} & $10^4$ samples & 0.48\% & 0.65\% & 1.7\% \\
                                   & $10^5$ samples & 0.15\% & 0.21\% & 0.52\% \\
                                   & $10^6$ samples & 0.05\% & 0.07\% & 0.18\% \\[5pt]
        \textbf{\textit{PyMC}} & $4 \times 5000$ samples & 0.09\% & 0.07\% & 0.07\% \\
        \bottomrule
    \end{tabular}
    \caption{\label{tab: mape large hier} Average percentage errors on the reconciled mean (Gaussian, large hierarchy).}
\end{table*}

\begin{table*}[h]
    \centering
    \begin{tabular}{c@{\hskip 1cm}ccc}
        \toprule
        {} & \multicolumn{3}{c}{Number of samples}\\
        \cmidrule{2-4}\\
        \addlinespace[-10pt]
        & $10^4$ & $10^5$ & $10^6$\\
        \midrule
        \textbf{IS} & 0.02 {\footnotesize $\pm$ 0.00} & 0.17 {\footnotesize $\pm$ 0.03} & 1.74 {\footnotesize $\pm$ 0.37} \\
        \textbf{BUIS}    & 0.10 {\footnotesize $\pm$ 0.01} & 0.90 {\footnotesize $\pm$ 0.15} & 13.1 {\footnotesize $\pm$ 2.19} \\
        \bottomrule
    \end{tabular}
    \caption{\label{tab: times large hier}Average computational times (Gaussian, large hierarchy). The average time for \textit{PyMC} ($4$ chains with $5,000$ samples each) is $102.56$ {\footnotesize $\pm$ 29.93}.}
\end{table*}

\section{Efficiency of IS}
\label{sec: app KL}

It is well-known that vanilla importance sampling is not effective to sample from high dimensional distributions; this prevents using it to reconcile large hierarchies.
We also obtain low performances when the proposal distribution $\nuhat_b$ is not a good approximation of the target distribution $\nutil$.
The following result relates the Kullback-Leibler divergence \citep{kullback1951information} between the base and reconciled distribution to the efficiency of IS.
\begin{proposition} \label{prop: kl}
Let $\BH$ be a random vector distributed as $\nuhat_b$, and let $W := \pihat(\Abf\BH,\BH) / \pihat_b(\BH)$.
Then, the Kullback-Leibler divergence of the base bottom distribution from the reconciled bottom distribution is given by
\begin{equation} \label{eq: KL base reconc}
\kl(\nuhat_b \,\|\, \nutil) = \log\left( \E[W] \right) - \E[\log(W)].
\end{equation}
\end{proposition}

\begin{proof}
First, we recall that, given a pair of absolutely continuous probability distributions $\mu$ and $\nu$, the Kullback-Leibler (KL) divergence is defined as
\[
\kl(\mu \,\|\, \nu) = \int \log\left(\frac{p(x)}{q(x)}\right) p(x) \, dx,
\]
where $p$ and $q$ are the densities of, respectively, $\mu$ and $\nu$.
The discrete case is completely analogous.

Now, let $\hat{\nu}_b$ be the base bottom forecast distribution, and $\nutil$ the reconciled distribution.
We recall that the density of $\nutil$ is given by 
\[\pitil(\bbf) = \frac{1}{c} \hat{\pi}(\Abf\bbf, \bbf),\] 
where 
\begin{align*}
c &:= \int \hat{\pi}(\Abf\bbf, \bbf) \, d\bbf \\ 
&= \int \frac{\hat{\pi}(\Abf\bbf, \bbf)}{\hat{\pi}_b(\bbf)} \hat{\pi}_b(\bbf) \, d\bbf \\
&= \E\left[\frac{\hat{\pi}(\Abf \BH, \BH)}{\hat{\pi}_b(\BH)}\right]
\end{align*}
is the normalizing constant, and $\BH \sim \nuhat_b$.
Then, we have
\begin{align}
&\kl(\hat{\nu}_b \,\|\, \nutil) 
= \int \log\left( c \, \frac{\hat{\pi}_b(\bbf)}{\hat{\pi}(\Abf \bbf, \bbf)} \right) \hat{\pi}_b(\bbf) \, d\bbf \nonumber \\
&= \log(c) - \int \log\left( \frac{\hat{\pi}(\Abf \bbf, \bbf)}{\hat{\pi}_b(\bbf)} \right) \hat{\pi}_b(\bbf) \, d\bbf \nonumber \\
&= \log\left( \E\left[\frac{\hat{\pi}(\Abf \BH, \BH)}{\hat{\pi}_b(\BH)}\right] \right) 
- \E\left[\log\left(\frac{\hat{\pi}(\Abf \BH, \BH)}{\hat{\pi}_b(\BH)}\right)\right] \nonumber \\
&= \log\left( \E[W] \right) - \E[\log(W)].  
\end{align}
\end{proof}

Note that the right-hand side of \eqref{eq: KL base reconc}
is a measure of the dispersion of the random variable $W$. 
Indeed, by the Jensen's inequality, it is always non-negative, and it is zero when $W$ is constant a.s.;
it gets larger as $W$ becomes more dispersed.
In the context of the measures of inequality, it usually referred to as Mean Logarithm Deviation \citep{haughton2009handbook}.
%
Moreover, from \eqref{eq: weights}, we have that the importance sampling weights are IID copies of $W$.
Hence, the more distant are the base and the reconciled distribution, in terms of Kullback-Leibler divergence, the more dispersed are the IS weights.
A large dispersion of the weights leads to a poor performance of importance sampling \citep{martino2017effective}.
As the incoherence level $\epsilon$ grows,
the distance between the distributions of $\Abf \BH$ and $\UH$ grows, and therefore also the distance between  $\nuhat_b$ and $\nutil$,
as the reconciled distribution merges the information coming from the bottom and the upper variables.

\end{document}